\documentclass[10pt,twocolumn,letterpaper]{article}

\usepackage{cvpr}              %
\usepackage{multirow}
\usepackage{comment}

\usepackage[table]{xcolor} 
\definecolor{best1}{RGB}{222,242,212}
\definecolor{best2}{RGB}{255,250,212}

\definecolor{cvprblue}{rgb}{0.21,0.49,0.74}
\usepackage[pagebackref,breaklinks,colorlinks,citecolor=cvprblue]{hyperref}

\def\paperID{213} %
\def\confName{CVPR}
\def\confYear{2024}

\title{VideoRF: Rendering Dynamic Radiance Fields as 2D Feature Video Streams}

\author{Liao Wang$^{1,3*}$
\and
Kaixin Yao$^{1,3*}$
\and
Chengcheng Guo$^{1}$
\and
Zhirui Zhang$^1$
\and
Qiang Hu$^{1}$
\and
Jingyi Yu$^{1}$
\qquad \qquad Lan Xu$^{1}$$^{\dagger}$  \qquad \qquad
Minye Wu$^{2}$$^{\dagger}$   \\
$^{1}$ ShanghaiTech University \qquad $^{2}$ KU Leuven \qquad $^{3}$ NeuDim \\
}

\begin{document}

\twocolumn[{
  \maketitle
  \vspace{-13mm}
  \begin{center}
  \includegraphics[width=0.99\textwidth]{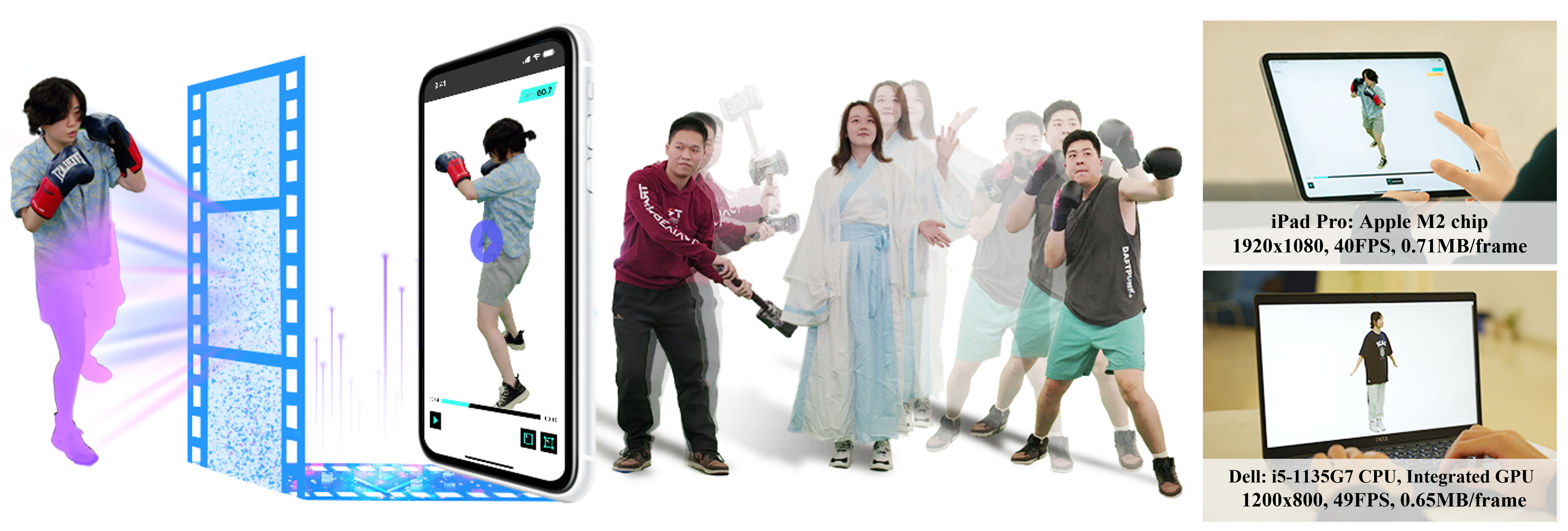}
  \vspace{-2ex}
  \captionof{figure}{\small{Our proposed VideoRF views dynamic radiance field as 2D feature video streams combined with deferred rendering.  This technique facilitates hardware video codec and shader-based rendering,
  enabling smooth high-quality rendering across diverse devices.}}
  \label{fig:teaser}
  \end{center}
}]
{\let\thefootnote\relax\footnote{* Authors contributed equally to this work.
${\dagger}$ The corresponding authors are Minye Wu
		(minye.wu@kuleuven.be) and Lan Xu (xulan1@shanghaitech.edu.cn).
}}\par
\begin{abstract}
Neural Radiance Fields (NeRFs)  excel in photorealistically rendering static scenes.
However, rendering dynamic, long-duration radiance fields on ubiquitous devices remains challenging, due to data storage and computational constraints.
In this paper, we introduce VideoRF, the first approach to enable real-time streaming and rendering of dynamic radiance fields on mobile platforms.
At the core is a serialized 2D feature image stream representing the 4D radiance field all in one.
We introduce a tailored training scheme directly applied to this 2D domain to impose the temporal and spatial redundancy of the feature image stream. 
By leveraging the redundancy, we show that the feature image stream can be efficiently compressed by 2D video codecs, which allows us to exploit video hardware accelerators to achieve real-time decoding. 
On the other hand, based on the feature image stream, we propose a novel rendering pipeline for VideoRF, which has specialized space mappings to query radiance properties efficiently. 
Paired with a deferred shading model, VideoRF has the capability of real-time rendering on mobile devices thanks to its efficiency. 
We have developed a real-time interactive player that enables online streaming and rendering of dynamic scenes, offering a seamless and immersive free-viewpoint experience across a range of devices, from desktops to mobile phones.
Our project page is available at \url{https://aoliao12138.github.io/VideoRF/}.

\end{abstract} 
\section{Introduction}
\label{sec:intro}
Photorealistic Free-Viewpoint Video (FVV) of dynamic scenes offers an immersive experience in virtual reality and telepresence. Work involving Neural Radiance Fields (NeRFs) has shown great potential in creating photorealistic Free-Viewpoint Videos (FVVs). However, there are still challenges in smoothly delivering and rendering FVVs using NeRFs on commonly used devices, similar to the ease of watching online videos.
The difficulty lies in reducing the data capacity for transmitting and storing long sequences and ensuring a low, mobile-compatible computational load.

Neural Radiance Field (NeRF) \cite{mildenhall2020nerf} surpasses traditional 3D reconstruction methods in photorealistic novel view synthesis. 
Several works extend NeRF to dynamic scenes by maintaining a canonical space and matching it implicitly \cite{park2020deformable,pumarola2021dnerf,du2021nerflow} or explicitly \cite{liu2022devrf} to align with each frame's live space. 
However, their dependence on canonical space limits effectiveness in sequences with large motions or topological changes. 
Other methods introduce new representations like 4D feature grids \cite{isik2023humanrf} or temporal voxel features \cite{TiNeuVox}, achieving impressive results on the scenes with topological transformations.
Yet, \cite{TiNeuVox} struggles with representing longer sequences owing to model capacity constraints, while \cite{isik2023humanrf} encounters streaming challenges due to large storage needs.
Recent efforts \cite{Wang_2023_rerf,song2023nerfplayer,li2022streaming} focus on compressing dynamic frames for streaming, but their computational intensity limits mobile applicability.
Concurrently, on mobile devices, real-time rendering of static scenes has been achieved by baking NeRF into mesh templates \cite{chen2022mobilenerf, bakedsdf, Rojas_2023_rerend} or texture assets \cite{Reiser2023SIGGRAPH}.
However, these techniques fall short for dynamic scenes as their per-frame representation becomes too bulky for real-time loading. Moreover, while NeRF compression methods \cite{Rho_2023_CVPR, takikawa2022variable, Wang_2023_rerf} can be employed, they introduce decoding or rendering overheads unsuitable for mobile platforms. Despite the existence of NeRF solutions for dynamic scenes and mobile optimization, a cohesive approach that effectively addresses both still poses a difficulty.

In this paper, we propose \textit{VideoRF} -- a novel neural modeling approach that enables real-time streaming and rendering of human-centric dynamic radiance fields on common mobile devices (see Fig.~\ref{fig:teaser}). 
Our key idea is to view the 4D feature volumes reconstructed from a dynamic scene as a 2D feature image stream, which is friendly to video codec.
Each feature image records the densities and appearance features of one frame.
We hence propose a rendering pipeline for this representation. 
In this rendering pipeline, VideoRF uses a set of mapping tables to connect the 3D space with the 2D feature images. This allows for O(1) density and feature retrieval from the images for every 3D sample point in the space.
To reduce the computational complexity of rendering, we adopt the deferred shading model~\cite{Reiser2023SIGGRAPH} with a global tiny MLP to decode the integrated feature vectors into pixel colors. 
All rendering operations here are low-cost and compatible with the fragment shader model, making it possible to implement this rendering pipeline on various devices that have GPUs.

Second, we present a sequential training scheme to effectively generate the 2D feature image stream. 
Specifically, VideoRF adaptively groups frames by analyzing motion from the sequential data to ensure temporal stability for mapping table generation. 
We deploy 3D and 2D Morton sorting techniques to improve spatial consistency in the mapping. 
VideoRF contains spatial and temporal continuity regularizers that are applied on the mapped 2D feature images using the mapping table. 
This training strategy enforces the temporal and spatial redundancy to the 2D feature image stream. 
We show that our feature image stream with spatiotemporal sparsity
can be efficiently compressed by off-the-shelf video codecs and reaches high compression rates. 

Moreover, based on these findings, we build a cross-platform player based on VideoRF that can play FVVs in real-time, supported by video hardware accelerators. 
In this way, users can interactively drag, rotate, pause, play, fast-forward, rewind, and jump to specific frames, providing a viewing experience as seamless as watching online videos on various devices, including smartphones, tablets, laptops, and desktops which was unseen before.

To summarize, our contributions include:
\begin{itemize}
	\item We propose VideoRF, a novel approach to enable real-time dynamic radiance field decoding, streaming and rendering on mobile devices.
 \item We present an efficient and compact representation, 
which represents 4D radiance field into 2D feature stream 
with low rendering complexity to support hardware video 
codec and shader rendering.
	
 \item We introduce a training scheme to directly impose spatial-temporal consistency on our 2D feature stream for efficient compression.

\end{itemize}
\section{Related work}
\label{sec:Relatedwork}
\textbf{Novel View Synthesis for Dynamic Scenes.} 
Dynamic scenes pose a greater challenge in achieving realistic view synthesis results due to moving objects. One approach is to reconstruct the dynamic scene and render the geometry from new viewpoints. The conventional RGB \cite{kim2010dynamic, collet2015high,ranftl2016dense,lv2018learning,li2019learning,luo2020consistent, zhao2022human} or RGB-D \cite{motion2fusion, TotalCapture, UnstructureLan,newcombe2015dynamicfusion, FlyFusion,jiang2022neuralfusion, 10204718} solutions have been extensively investigated. 
Mesh-based neural network 
 \cite{wood2000surface,waechter2014let,chen2018deep} techniques are effective for compact data storage and can record view-dependent texture \cite{wood2000surface,chen2018deep}, but such methods heavily rely on geometry, especially perform poorly for topologically complex scenarios.

Many methods extend NeRF into the dynamic view synthesis settings. 
Some methods \cite{pumarola2021dnerf,li2020neural,xian2020space,st-nerf,Gao-ICCV-DynNeRF,du2021nerflow,park2021hypernerf, wang2021ibutter,li2022dynerf, lin2022efficient, yu2023dylin, li2023dynibar} handle spatial change directly conditions on time and \cite{peng2021neural, pengtpami, Gafni_2021_CVPR, xu2023latentavatar, zheng2023avatarrex} use feature latent code to represent time information. 
However, they do not support the streaming of long sequences due to the limited representation. 
Others learn spatial offsets from the live scene to the canonical radiance field by using explicit voxel matching \cite{liu2022devrf}, skeletal poses \cite{luo2022artemis, peng2021animatable, kwon2023deliffas}, deformed hashtable \cite{kirschstein2023nersemble} and deformed volume \cite{kappel2023fast, tretschk2020non, zhao2023havatar, xu2023avatarmav, tretschk2024scenerflow, neus2}. 
While these methods can successfully render dynamic scenes with photo-realistic quality,  their heavy reliance on the canonical space makes these methods vulnerable to long sequences, large motions, and topological alterations.
Recently, several methods using 4D planes  \cite{kplanes_2023,shao2023tensor4d,Cao2022hexplane,isik2023humanrf,xu20234k4d}, voxel grid \cite{TiNeuVox}, Fourier representation \cite{wang2022fourier}, residual layers \cite{ResFields2023}  or dynamic MLP maps \cite{peng2023representing} to represent dynamic scene. However, the inference computational complexity makes it difficult for them to implement streaming and decoding on mobile devices.
Additionally, several methods such as 
 \cite{song2023nerfplayer, Wang_2023_rerf, li2022streaming} have managed to facilitate the streaming of dynamic radiance fields by reducing the capacity of each frame, but the lack of general hardware acceleration makes it challenging to represent dynamic scenes in real-time on mobile phones.
\vspace{1ex}
\\
\textbf{Cross Device Neural Radiance Field Rendering.} %
Recent works have demonstrated that static neural radiance fields can be rendered on mobile devices. They achieve this by directly converting the neural radiance field to mesh \cite{tang2022nerf2mesh} or using traditional texture to represent the scene. \cite{chen2022mobilenerf, Neural-Enhanced, bakedsdf}  have endeavored to augment the capabilities of neural rendering by employing mesh templates and feature texture, \cite{Rojas_2023_rerend, bozic2022neural} facilitating the rendering process through fragment shader to achieve surface-like rendering with similar ideas.
\cite{Wan_2023_CVPR} represents neural radiance features encoded on a two-layer duplex mesh for better rendering quality.

Also, technologies \cite{Reiser2023SIGGRAPH, Cao2022hexplane, cao2023realtime} embody the principle of utilizing implicit neural representations to handle radiance rendering on mobiles. Although all these methods are fairly effective at compressing static scenes and enabling rendering on mobile devices, they still do not support streaming for dynamic scenes, particularly in the case of long sequences of dynamic scenes, they will still occupy a large amount of storage space.
\begin{figure}[t]
\begin{center}
    \includegraphics[width=\linewidth]{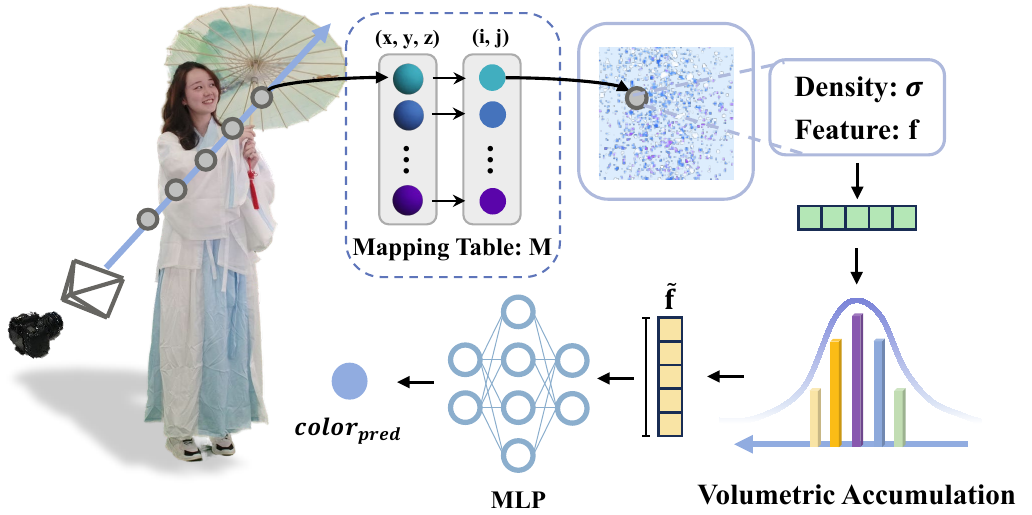}
\end{center}
\vspace {-5mm}
\caption{Demonstration of our rendering. For each 3D sample point, its density $\sigma$ and feature $\mathbf{f}$ are fetched from the 2D feature image through the mapping table $\mathbf{M}$. Each point feature is first volumetrically accumulated to get the ray feature $\tilde{\mathbf{f}}$ and pass MLP $\Phi$ to decode the ray color. }
\label{fig:pipeline}
\vspace {-5mm}
\end{figure}
\vspace{1ex}
\\
\textbf{NeRF Acceleration and Compression.}
NeRF demonstrates exceptional performance in generating images from arbitrary viewpoints, but it suffers from slow rendering efficiency. Some methods focus on integrating a compact structure with a simplified MLP, which reduces the complexity of MLP calculations in traditional NeRFs. Key strategies have been explored, including the employment of voxel grid \cite{sun2021direct,li2022streaming}, octrees \cite{yu2021plenoctrees,fridovich2022plenoxels, wang2022fourier}, tri-planes \cite{EG3D}, hashing encoding \cite{muller2022instant}, codebook \cite{takikawa2022variable, li2023compact}, tensor decomposition \cite{chen2022tensorf,tang2022compressible} to accomplish this. Using explicit structure makes training and rendering faster, but it also means that the 3D structure takes up more storage space. 

Therefore, while methods such as CP-decomposition \cite{chen2022tensorf}, rank reduction \cite{tang2022compressible}, and vector quantization \cite{takikawa2022variable} manage to attain modest levels of data compression, their application remains confined to static scenes.
Additionally, some approaches \cite{Wang_2023_rerf, Rho_2023_CVPR, Deng_2023_WACV} opt for post-processing techniques, initially utilizing baseline methods to train a decent NeRF representation, then special encoding and decoding schemes are employed to reduce the storage space. However, these kinds of methods are not hardware-friendly, the inference process is computationally demanding and cannot support real-time streaming on mobile devices. In contrast, our VideoRF uses codec tailored training, hardware-friendly decoding and shader rendering, which enables real-time dynamic radiance field for long sequences with large motion.

\section{VideoRF Representation} \label{sec:rep}
To facilitate dynamic radiance field rendering on mobile devices, 
we adopt a representation that aligns with the video codec format and maintains low computational complexity for shader-based rendering.
We propose turning the 3D volume representation into a 2D formulation,
 coupled with a highly efficient rendering pipeline.%

\label{sec:3.1}
In our approach, each frame of the radiance field is represented as a feature image $\mathbf{I}$ where the first channel stores density and the remaining $h$ channels store feature.
As depicted in Fig. \ref{fig:pipeline}, given a 3D vertex position $\mathbf{x}$, we retrieve its density $\sigma$ and feature $\mathbf{f}$ using the equation:
\begin{equation}
\sigma , \mathbf{f} = \mathbf{I}[\mathbf{M}(\mathbf{x})],
\label{eq:fetch}
\end{equation}
where $\mathbf{M}$ is the 3D-to-2D mapping table from Sec. \ref{sec:3.2}.
This mapping table not only effectively excludes empty space to reduce storage but also specifics mapping from each non-empty 3D vertex to a corresponding 2D pixel. The lookup operation is highly efficient, with a time complexity of O(1), facilitating rapid and convenient queries.
It's worth noting that our sequence of feature images is in a 2D format, which is friendly to the video codec hardware. %

\begin{figure*}[t]
\vspace {-3mm}
\begin{center}
    \includegraphics[width=\linewidth]{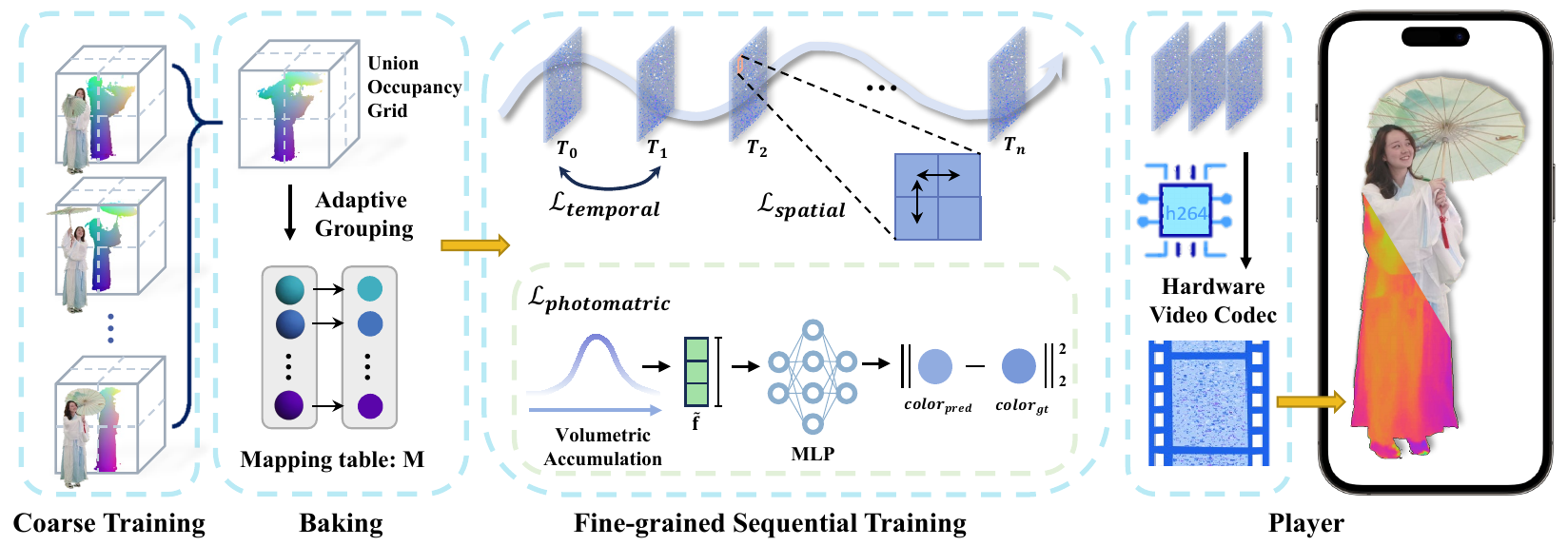} 
\end{center}
\vspace {-5mm}
\caption{Overview of our video codec-friendly training. First, we apply our grid-based coarse training \cite{sun2021direct} to generate per-frame occupancy grid  $\mathbf{O}^{t}$. Then, during baking, we adaptively group each frame and create a mapping table $\mathbf{M}$ for each group. Next, we sequentially train each feature image $\mathbf{I}^t$ through our spatial, temporal and photometric loss. Finally, feature images are compressed into the feature video streaming to the player.}
\label{training}
\vspace{-5mm}
\end{figure*}
For rendering, %
 inspired by \cite{Reiser2023SIGGRAPH, hedman2021snerg},  we use a deferred rendering model. We first accumulate the features along the ray:
\begin{equation}
\label{equ:volumerendering}
\begin{split}
\tilde{\mathbf{f}}(\mathbf{r}) &=\sum_{k=1}^{n_s} T_k(1-\exp(-\sigma_k\delta_k))\mathbf{f}_k, \\
T_k &= \operatorname{exp}\left(-\sum_{j=1}^{k-1}\sigma_j\delta_j\right),
\end{split}
\vspace{-20pt}
\end{equation}
where $n_s$ is the number of sample points along the ray $\mathbf{r}$. $\sigma_k$, $\mathbf{f}_k$, $\delta_k$ denotes the density, feature of the samples, and the interval between adjacent samples respectively. The view-dependent color of the ray is then computed using a tiny global MLP $\Phi$ shared across the frames as:
\begin{equation}
\tilde{C}(\mathbf{r})= \operatorname{sigmoid} \left( \Phi\left( \tilde{\mathbf{f}}(\mathbf{r}), \mathbf{d}\right) \right),
\end{equation}
where $ \mathbf{d}$ is the view direction of the ray after positional encoding \cite{mildenhall2020nerf}. In this way, the computational burden is significantly reduced as each ray requires only a single MLP decoding
which can be implemented in a shader for real-time rendering on mobile devices.

\section{Video Codec-friendly Training}  \label{sec:training}
In this section, we propose a training scheme, as depicted in Fig. \ref{training}, to achieve a high compression rate by maintaining spatial and temporal consistency. 
For spatial aspect,
we incorporate 3D-2D Morton sorting to preserve 3D continuity and apply a spatial consistency loss directly on 2D feature images to enforce spatial coherence. 
On the temporal front, we employ adaptive grouping, which allows frames within a group to share mappings, thereby reducing temporal disruptions, and temporal consistency loss, which serves to further reinforce temporal coherence.
These consistencies are crucial which provide a stable and predictable layout for video codecs. 

\subsection{Baking and Map Generation}  \label{sec:3.2}
\begin{figure}[t]
	\begin{center}
		\includegraphics[width=1.0\linewidth]{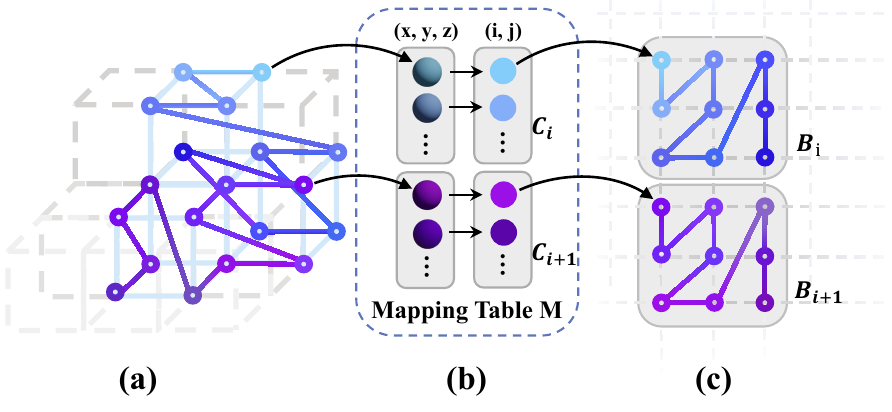} %
	\end{center}
 \vspace{-6mm}
	\caption{Illustration of our mapping table generation. We first perform 3D Morton sorting \textbf{(a)} on each nonempty vertice and group it into chunks $C_i$ \textbf{(b)}. Next, we lay out each chunk into each block $B_i$ of the feature image, arranged in 2D Morton order \textbf{(c)} within it.  }  
	\label{baking}
  \vspace{-5mm}
\end{figure}

\textbf{Coarse stage pre-training.}
Given multiview images of each frame, we first adopt an off-the-shelf approach \cite{sun2021direct} to generate the explicit density grid $\mathbf{V}_{\sigma}^{t}$ for each frame $t$ independently.
We then create a per-frame occupancy grid $\mathbf{O}^{t}$ by masking in voxels whose opacity exceeds a certain threshold $\gamma$.
This coarse stage sets the foundation for our subsequent adaptive grouping in the baking stage.
\vspace{1ex}
\\
\textbf{Adaptive group.} \label{sec:group}
Generating an independent mapping table for each frame $t$ could disrupt temporal continuity, leading to increased storage requirements after video codec compression. This issue arises because the same 3D point in adjacent frames might map to vastly different 2D positions on the feature image. On the other hand, uniformly applying a single mapping table across all frames could introduce significant spatial redundancy. Considering that a 3D point may only be occupied at sparse intervals, it could remain underutilized for most of the duration, resulting in inefficiency.
To balance these factors, we divide the sequence into Groups of Frames (GOFs), maintaining a fixed resolution for our 2D feature image and adaptively determining the number of frames in each group. 
For a set of consecutive frames $\left\{i, i+1, \ldots, i+n\right\}$, 
we identify the maximum frame number $\alpha$ such that the number of occupied voxels in the union from $i$ to $\alpha$ does not exceed our pixel limit $\theta$:
\begin{equation}
\operatorname{argmax}_\alpha  \operatorname{g}\left(\bigcup_{j=i}^\alpha \mathbf{O}^{j} \right) \leqslant \theta ,
\end{equation}
where $\operatorname{g}()$ means the number of occupied grids in the union occupancy grid. Then, we will set the frame
$i$ to frame $\alpha$ as a GOF and frame $\alpha+1$ to be the start frame of
a new GOF.
In this way, all the frames in the group share the same mapping. 
The same 3D position within the group will be mapped to the same 2D pixel location which keeps temporal consistency and saves the storage. 
\vspace{1ex} \\
\textbf{3D Morton sorting.}
After the union occupancy grid is achieved, we apply Morton sorting to record its 3D spatial continuity.
Morton ordering, or Z-ordering, interleaves the binary representations of spatial coordinates, ensuring that spatially proximate entities remain adjacent in linear space.
As shown in Fig. \ref{baking}(a), we apply 3D Morton sorting to the vertices based on their position coordinates of the union occupancy grid. Vertices with a density below the threshold $\gamma$ are excluded. This process effectively maintains 3D spatial consistency in the ordering.
\vspace{1ex} \\
\textbf{2D block partitioning and 2D Morton sorting.}
To preserve the 3D spatial continuity within a 2D framework, we employ 2D Morton sorting and partition the feature images into blocks. 
This approach aligns with the block-wise compression of frames in video codecs, where blocks with local smoothness lead to more efficient storage.
 Specifically, we first divide the feature image into $N$ $8\times 8$ blocks, denoted as $B_i$, and correspondingly group the sorted vertices into $N$ $8\times 8$ chunks, denoted as $C_i$, as illustrated in Fig. \ref{baking}(b).
For each pixel $\mathbf{p}$ in a block $B_i$, we sort its relative position $(u,v)$  in 2D Morton order. As shown in Fig. \ref{baking}(c), each chunk $C_i$ is then mapped to a block $B_i$, arranged in 2D Morton order within the block, to form the mapping table $\mathbf{M}$. This 2D Morton ordering ensures that sorted values in the 3D Morton ordering are positioned closely within each feature image block, facilitating efficient compression during the transformation process.

\subsection{Fine-grained Sequential Training}
With the aid of our mapping table, we can sequentially train our feature images within each group through spatial consistency loss applied to 2D feature images and temporal consistency loss between frames.
\vspace{1ex} \\
\textbf{Spatial consistency loss.}
In video encoding, regions with homogeneous characteristics demonstrate high compression efficiency. This efficiency stems from the minimal variation in pixel values within these areas, leading to a significant reduction in high-frequency components post Discrete Cosine Transform (DCT) within the video codec. Consequently, these regions retain a higher proportion of low-frequency components. Furthermore, the quantized coefficients in such homogeneous regions are more likely to contain a greater number of zero values, facilitating a more compact data representation and reducing the overall data volume.
In order to enhance the homogeneity of our 2D feature image, we introduce a total variance loss, $\mathcal{L}_{\mathrm{spatial}}$, during our fine-grained training stage. 
For each channel of the feature image $\mathbf{I}$, we enforce its local smoothness by:
\begin{equation}
\mathcal{L}_{\mathrm{spatial}}=\frac{1}{|\mathcal{P}|} \sum_{\substack{\mathbf{p} \in \mathcal{V}}} \left( \Delta_u(\mathbf{p})+\Delta_v(\mathbf{p}) \right),
\end{equation}
where $\mathcal{P}$ is the pixels of the feature image,  $\Delta_u(\mathbf{p})$ shorthand for Manhattan distance between the feature value at pixel $\mathbf{p}:=(u,v)$ and the feature value at pixel $(u+1,v)$ normalized by the resolution, and analogously
for $\Delta_v(\mathbf{p})$.
By increasing the spatial sparsity, the storage of feature videos after video encoding is decreased at the same quality.
\vspace{1ex} \\
\textbf{Temporal consistency loss.} 
A naive per-frame training scheme will disrupt temporal continuities by failing to incorporate inter-frame information and resulting in a high bitrate. This is because the residuals between frames are stored after entropy encoding. 
To optimize storage efficiency, we focus on minimizing the differences in the feature space between the adjacent frames to reduce the entropy. 
During our sequential training, we enhance inter-frame similarities by regularizing the current feature image with its predecessor, except for the initial frame of each adaptive group. 
This is achieved by applying
\begin{equation}
\mathcal{L}_{\mathrm{temporal}}= \|\mathbf{I}^t-\mathbf{I}^{t-1}\|_1 ,
\end{equation}
for each frame $t$ in the group, ensuring small residuals between consecutive feature images.
By employing temporal smoothness in this manner, we can further mitigate entropy throughout the video codec process, thereby facilitating storage conservation.
\vspace{1ex} \\
\textbf{Training objective.} Our total loss function is formulated as:
\begin{equation}
\mathcal{L}_{\text {total }}= \mathcal{L}_{\mathrm{rgb}}+\lambda_s \mathcal{L}_{\mathrm{spatial}}  +\lambda_t \mathcal{L}_{\mathrm{temporal}}, 
\end{equation}
where $\lambda_s$ and $\lambda_t$ are the weights for our regular terms and $ \mathcal{L}_{\mathrm{rgb}}$ is the photometric loss,
\begin{equation}
\mathcal{L}_{\text {rgb}}=\sum_{\mathbf{r} \in \mathcal{R}}\|\mathbf{c}(\mathbf{r})-\hat{\mathbf{c}}(\mathbf{r})\|^2,
\end{equation}
where $\mathcal{R}$ is the set of training pixel rays; $\mathbf{c}(\mathbf{r})$ and $\mathbf{\hat c}(\mathbf{r})$ are the ground truth color and predicted color of a ray $\mathbf{r}$ respectively.

\begin{figure*}[t]
	\vspace{-0.5cm}
	\begin{center}
		\includegraphics[width=\linewidth]{ 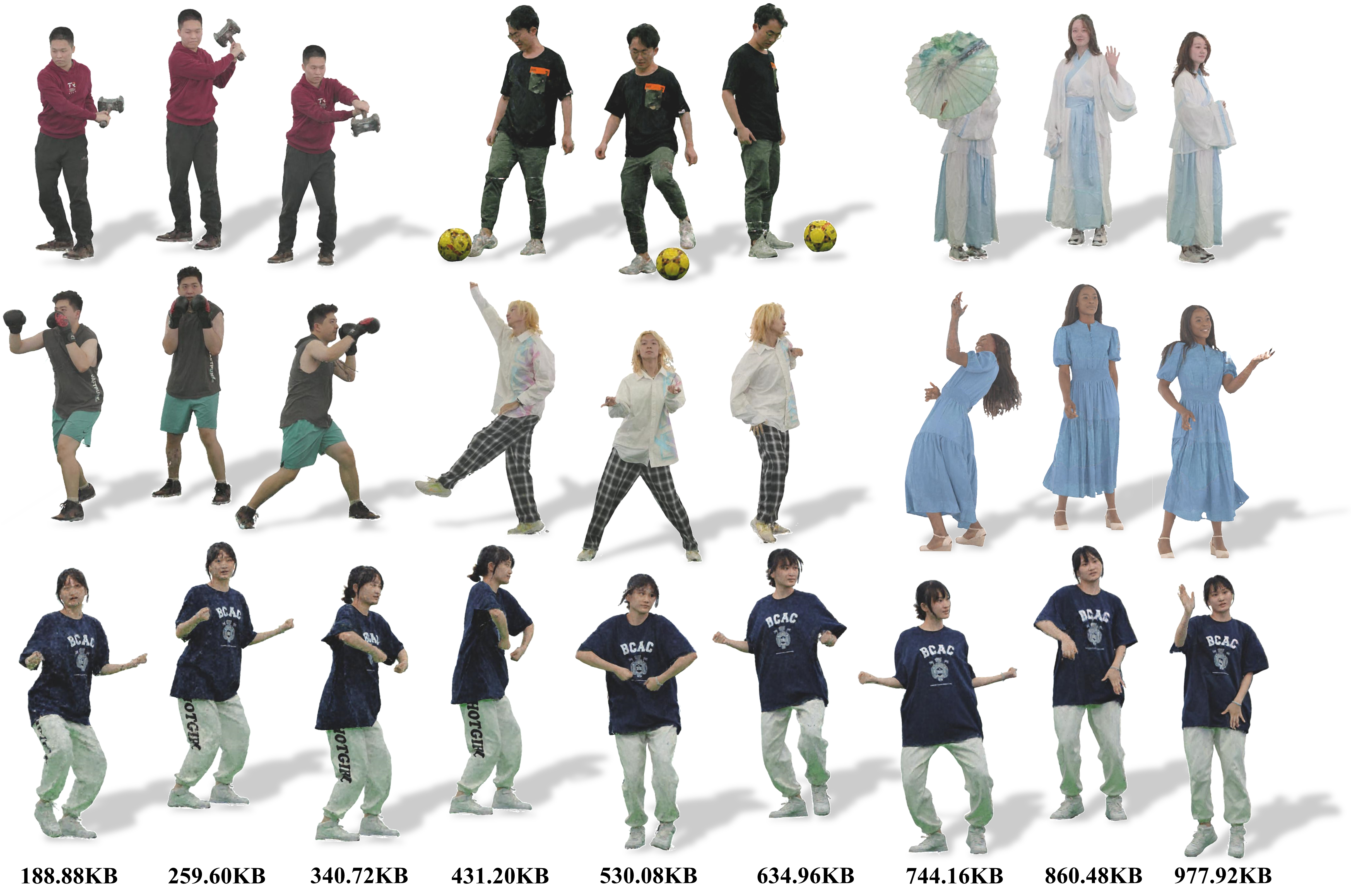}
	\end{center}
	\vspace{-0.6cm}
	\caption{Our VideoRF method generates results for inward-facing, 360$^\circ$~ video sequences featuring human-object interactions with large motion. The images in the last row illustrate our ability to implement variable bitrate in these sequences.}
	\label{fig:gallery}
	\vspace{-5mm}
\end{figure*}

\subsection{VideoRF Player} \label{player}
Finally, we implement a companion VideoRF player to stream and render dynamic radiance fields on mobile devices. 
Given our feature stream, we quantize it into uint8 format and use H.264 for video codec.
Note that the total resolution of feature images is under a 4K color image, ensuring seamless real-time streaming and decoding even on mobile devices.

For the rendering part, we implement it via a fragment shader. 
For fast raymarching, we employ a multi-resolution hierarchy
of occupancy grids for each group to skip empty space at different levels.
We leverage matrix multiplication in the shader to simulate our tiny MLP calculation.
Different from \cite{Reiser2023SIGGRAPH}, our method stores the mapping table in a 2D-to-3D format as an RGB image to save storage redundancy.
 Meanwhile, to speed up querying values for each sampled point, we adopt a parallel approach to expand a 2D feature image into a 3D volume in the player. 
 These techniques increase the overall speed while ensuring compact storage. 

The VideoRF player marks a significant milestone, as it first enables users to experience real-time rendering of dynamic radiance fields of any length. 
Within this player, users can drag, rotate, pause, play, fast forward/backward, seek dynamic scenes, or switch between different resolutions like watching online video, offering an
extraordinary high-quality free-viewpoint viewing experience.
This capability also extends across a wide range of devices, from smartphones and tablets to laptops and desktops, broadening the accessibility and applicability of dynamic radiance fields.

\section{Experimental Results}
In this section, we evaluate our VideoRF on a variety of challenging scenarios. We use the PyTorch Framework to train the model on a single NVIDIA GeForce RTX3090.
Our new captured dynamic datasets contain around 80 views at the resolution of 1920 ${\times}$ 1080 at 30 fps. To ensure the robustness of the algorithm, we also use the ReRF dataset and the HumanRF dataset for result demonstration in Fig. \ref{fig:gallery}. We can render images for immersive, 360$^\circ$ video sequences that capture human interactions with objects, especially for large motion and long duration sequences. Please refer to the supplementary video for more video results.
\begin{figure*}[t]
\vspace{-0.5cm}
\begin{center}
    \includegraphics[width=\linewidth]{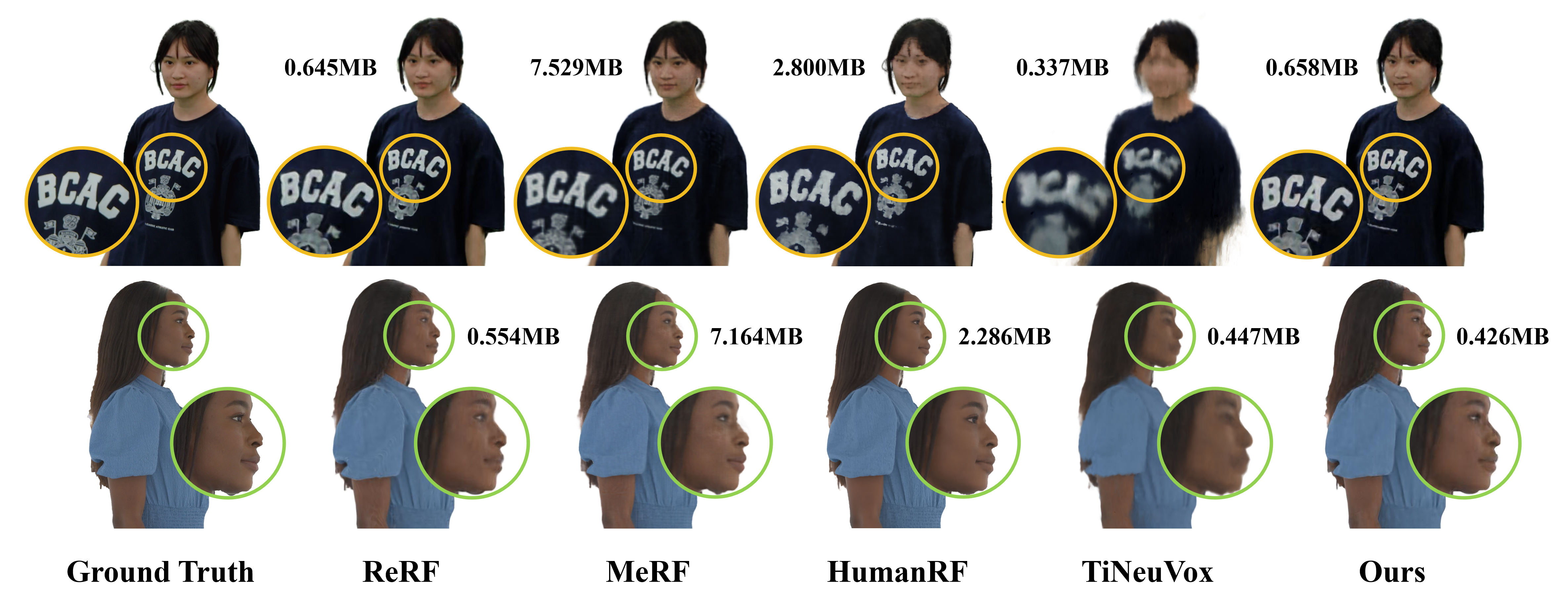}
\end{center}
\vspace{-0.7cm}
\caption{Qualitative comparison against dynamic scene reconstruction methods and per frame static reconstruction methods.
}
\vspace{-0.5cm}
\label{fig:comparison}
\end{figure*}

\subsection{Comparison}
To the best of our knowledge, our approach is the first real-time dynamic radiance field approach that can decode and render on mobile devices. Therefore, we compare to existing dynamic neural rendering methods, including ReRF \cite{Wang_2023_rerf}, HumanRF \cite{isik2023humanrf}, TiNeuVox \cite{TiNeuVox}  and per-frame static reconstruction methods like MeRF \cite{Reiser2023SIGGRAPH}.
We use ReRF \cite{Wang_2023_rerf} dataset and HumanRF \cite{isik2023humanrf} Actor3 sequence 1 for a fair comparison, both in storage memory and image quality. 
As shown in Fig. \ref{fig:comparison}, though TiNeuVox \cite{TiNeuVox} has small storage, it faces the challenge of intensifying blurring effects as the frame count rises. 
MeRF \cite{Reiser2023SIGGRAPH} suffers from the large storage which is not suitable for dynamic scenes when streaming and decoding for playing.
HumanRF \cite{isik2023humanrf}  though capable of representing dynamic scenes, also faces difficulties in streaming and rendering on mobiles due to its computation and storage load.
\begin{table}[t]
\begin{center}
\footnotesize
\centering\setlength{\tabcolsep}{6pt}
\renewcommand{\arraystretch}{1.1}
\setlength{\tabcolsep}{1.5mm}{\begin{tabular}{l | c | cccc }
\multicolumn{6}{c}{ \colorbox{best1}{best} \colorbox{best2}{second-best} } \\
\hline
 Dataset & Method & PSNR$\uparrow$ & SSIM$\uparrow$ & LPIPS $\downarrow$ & Size(MB)$\downarrow$ \\ 
\hline
\multirow{5}{*}{ReRF}
 & ReRF \cite{Wang_2023_rerf} & \cellcolor{best2}31.84 &  0.974 & 0.042 & \cellcolor{best2}0.645   \\
 & MeRF \cite{Reiser2023SIGGRAPH}    & 31.12 & \cellcolor{best2}0.975 & \cellcolor{best2}0.030  & 7.529  \\
 & HumanRF \cite{isik2023humanrf}    & 28.82 & 0.900 & 0.069 & 2.800     \\
 & TiNeuVox\cite{TiNeuVox}           & 22.70 & 0.923 & 0.083 & \cellcolor{best1}0.337  \\ \cline{2-6}
 & Ours                              & \cellcolor{best1}32.01 & \cellcolor{best1}0.976 &  \cellcolor{best1}0.023 & 0.658 \\ 
\hline
\multirow{5}{*}{Actors-HQ}
 & ReRF \cite{Wang_2023_rerf}        & 28.33 & 0.836 & 0.296 & 0.554    \\
 & MeRF \cite{Reiser2023SIGGRAPH}    & 27.22 & 0.807 & \cellcolor{best2}0.271 & 7.164     \\
 & HumanRF \cite{isik2023humanrf}    & \cellcolor{best1}28.98 &  \cellcolor{best1}0.888 & \cellcolor{best1}0.151 & 2.286      \\
 & TiNeuVox \cite{TiNeuVox}          & 22.98 & 0.752 & 0.430 &  \cellcolor{best2}0.447 \\ \cline{2-6}
 & Ours                              &  \cellcolor{best2}28.46 & \cellcolor{best2}0.838 & 0.278 & \cellcolor{best1}0.426  \\ 
\hline 
\bottomrule
\end{tabular}
}
\end{center}
\vspace{-5mm}
\caption{Qualitative comparison against dynamic scene reconstruction methods and per frame static reconstruction methods. 
}
\label{tab:comparison}
\vspace{-5mm}
\end{table}

We also conduct a quantitative comparison using metrics such as the peak signal-to-noise ratio (PSNR), structural similarity index (SSIM), and the Learned Perceptual Image Patch Similarity (LPIPS) as metrics shown in Tab. \ref{tab:comparison}.
For a fair comparison in our experiments, in the ReRF dataset, we test the first 200 frames on the scene Kpop. We use 6 and 39 as test views and the others as training views. As for the HumanRF dataset,  we apply the test methods as referenced in HumanRF on the first 250 frames on Actor3, sequence 1. We utilize the values reported for HumanRF and TiNeuVox methods directly from the HumanRF study.
In the ReRF dataset, which includes large motion scenarios, we match the rendering quality of the standard ReRF, while also maintaining a compact data storage size. Meanwhile, in the HumanRF dataset with relatively small motion, we achieve a rendering quality that is second only to HumanRF, accomplished with the most efficient storage utilization.
It's worth noting that, our method is the only method that can provide both mobile and dynamic rendering tasks. 
The existing dynamic reconstruction methods like TiNeuVox \cite{TiNeuVox}, HumanRF \cite{isik2023humanrf} and ReRF \cite{Wang_2023_rerf} all cannot effectively utilize hardware encoding and decoding techniques, preventing them from being displayed on mobile devices, especially for the dynamic scene of a long sequence.

\begin{figure}[t]
	\begin{center}
		\includegraphics[width=0.98\linewidth]{ 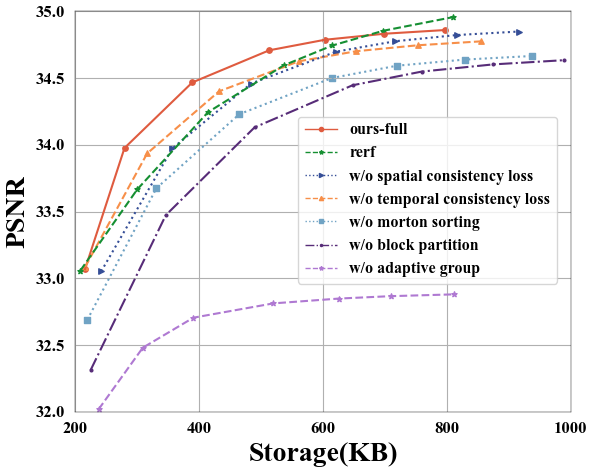}
	\end{center}
	\vspace{-0.7cm}
	\caption{\textbf{Rate distortion curve.} The rate distortion curve illustrates the efficiency of various components within our system. We use different quantization factors to obtain the average rendering quality at different capacities. Our full model stands out as the most compact, allowing for flexible bitrate adjustments to meet diverse storage needs.}
	\label{fig:eval1}
	\vspace{-5mm}
\end{figure}

\begin{figure*}[t]
	\begin{center}
	\includegraphics[width=0.98\linewidth]{ 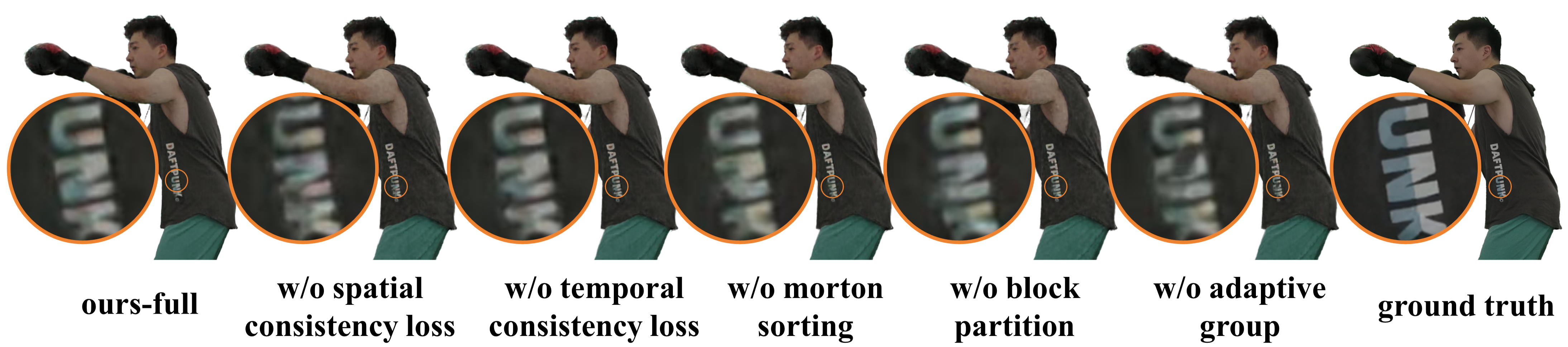}
	\end{center}
	\vspace{-0.5cm}
	\caption{Qualitative evaluation of different variations in our method at 600KB.}
	\label{fig:eval2}
	\vspace{-5mm}
\end{figure*}

\subsection{Evaluation}
\textbf{Ablation study.}
We analyze the impact of spatial consistency, temporal consistency, Morton sorting, block storage, and adaptive grouping on compression and rendering. 
In the case of models that do not employ 3D and 2D Morton sorting, spatial point data is sorted sequentially in row-major order while employing 2D block-wise storage. 
For models not utilizing block storage, a 3D spatial Morton sorting is applied, followed by storage in a row-major format within the 2D feature space. 
For models without adaptive grouping,
 the mapping table for each frame is calculated only based on the occupancy grid of the current frame.
As illustrated in Fig. \ref{fig:eval1}, our full model exhibits the best performance in terms of rendering quality and storage efficiency. 
This demonstrates the effectiveness of our modules in reducing capacity by maintaining consistency across time and space.
While ReRF \cite{Wang_2023_rerf} exhibits slightly better quality in the storage range of 700-800KB, our method consistently demonstrates higher PSNR across most capacity ranges and supports mobile rendering. 
Fig. \ref{fig:eval2} shows that under a 600KB storage limit, our complete model yields more realistic results with less compression blur.
\vspace{1ex}
\\
\textbf{Cross device runtime analysis.} 
We evaluate the runtime breakdown analysis of VideoRF as detailed in Tab. \ref{tab:devices}. Our experimental setup includes a Desktop with an i7-12700F CPU and NVIDIA RTX3090 GPU, a Laptop with an i5-1135G7 CPU and Integrated GPU, a Tablet (iPad Pro) with an M2 chip, and a phone (iPhone 14 Pro) with an A16 Bionic chip. For Desktops and Laptops, we employ a Python code base along with ModernGL. Meanwhile, for Tablet and Smartphone, the VideoRF player is developed using Swift and Metal. 
It's worth noting that the video decoding part primarily relies on CPU performance, while the rendering part mainly depends on GPU performance. These two parts operate asynchronously and simultaneously. The other part mainly covers operations such as data conversion between the GPU and CPU. Our results demonstrate that VideoRF allows users to enjoy free-view videos at high frame rates on multiple devices, providing an experience that rivals the smoothness of watching 2D videos on platforms such as YouTube.
\begin{table}[t]
\begin{center}
\small
\centering\setlength{\tabcolsep}{6pt}
\renewcommand{\arraystretch}{1.1}
\setlength{\tabcolsep}{1.5mm}{\begin{tabular}{l | ccccc }
\hline
Devices  & FPS & Video Decoding & Rendering & Others \\ 
\hline
Desktop   & 116 & 5.548 ms  & 3.602 ms & 3.072 ms\\  %
Laptop   & 25 & 10.54 ms & 29.31 ms & 11.69 ms\\  %
Tablet   & 40 & 3.822 ms & 23.86 ms & 1.150 ms\\ %
Phone  & 23 & 13.38 ms & 40.82 ms & 3.003 ms\\ %
\hline 
\bottomrule
\end{tabular}
}

\end{center}
\vspace{-5mm}
\caption{Runtime analysis across different devices when processing the same HD image with a resolution of 1920 $\times$ 1080.}
\label{tab:devices}
\vspace{-4mm}
\end{table}
\vspace{1ex}
\\
\textbf{Storage of different components analysis.} 
We present the storage requirements of each VideoRF component in Tab.~\ref{tab:storage}. This encompasses the average file sizes for several key elements: the feature images for model detail, 3D-to-2D mapping table, occupancy images used to efficiently skip over empty spaces, and MLP parameters for the neural network. Note that our model's total average size is a mere 669.78KB. This compact representation facilitates rapid streaming across various devices.

\begin{table}[t]
\begin{center}
\small
\centering\setlength{\tabcolsep}{6pt}
\renewcommand{\arraystretch}{1.1}
\setlength{\tabcolsep}{1.5mm}{\begin{tabular}{l | ccccc } 

\hline
        Components & Size(KB) \\
        \hline
		Feature Images & 661.62 \\
		3D to 2D Mapping Table & 2.58 \\
		Occupancy Images & 2.18 \\ 
        MLP Parameters & 3.40 \\ \hline
        Total Size & 669.78 \\
\hline 
\bottomrule
\end{tabular}
}

\end{center}
\vspace{-5mm}
\caption{Storage of different components. The result is averaged over a sequence of Kpop scene from ReRF \cite{Wang_2023_rerf} dataset.}
\label{tab:storage}
\vspace{-4mm}
\end{table}
\section{Discussion}
\textbf{Limitation.} 
As the first trial to enable a real-time dynamic radiance field approach capable of decoding and rendering on mobile devices, our approach presents certain limitations. First, high-quality reconstruction requires a complex system that captures multiple views, which is notably expensive. Second, our training time increases with the number of frames, especially for long sequences. Additionally, our method currently lacks support for temporal interpolation, signifying a direction for future exploration.
\vspace{1ex}
\\
\textbf{Conclusion.} 
We have presented VideoRF, a novel approach enabling real-time streaming and rendering of dynamic radiance fields on mobile devices. 
Our VideoRF innovatively processes feature volumes as 2D feature streams and adopts deferred rendering to effectively leverage classical video codecs and rasterization pipelines.
Our video codec friendly training scheme is implemented on 2D feature space to enhance spatial-temporal consistency for compactness.
Additionally, our tailored player supports seamless streaming and rendering of dynamic radiance fields across a range of devices, from desktops to smartphones.
Our experiments demonstrate its capability for compact and effective dynamic scene modeling.
With the unique ability of real-time rendering of dynamic radiance fields on mobile devices, we believe that our approach marks a significant step forward in neural scene modeling and immersive VR/AR applications.

{
    \small
    \bibliographystyle{ieeenat_fullname}
    \bibliography{main}
}

\clearpage
\setcounter{page}{1}
\maketitlesupplementary

\section{Training Details for VideoRF}
\label{sec:training}
\subsection{Coarse Stage Pre-training and Baking}
Given a long-duration multi-view sequence, we initially adopt the approach from DVGO~\cite{sun2021direct} to generate an explicit density volume grid $\mathbf{V}_{\sigma}$  and color feature grids $\mathbf{V}_{c}$ representation for each frame $t$. 
Following ReRF~\cite{Wang_2023_rerf}, we employ a global MLP $\Phi_c$  during this coarse stage training. 
This MLP comprises a single hidden layer with 129 channels, and we set the color feature dimension at $h=12$.
Throughout the training, we incrementally upscale the volume grid, from  $(125\times125\times 125)\rightarrow (150\times150\times 150)\rightarrow  (200\times200\times 200)\rightarrow (250\times250\times 250)$, after reaching the training step 2000, 4000 and 6000, respectively.
For loss calculation, we utilize both the photometric MSE loss and the total variation loss on  $\mathbf{V}_{\sigma}$, expressed as:
\begin{equation}
\mathcal{L}_{\text {rgb}_\text {{coarse}}}=\sum_{\mathbf{r} \in \mathcal{R}}\|\mathbf{c}(\mathbf{r})-\hat{\mathbf{c}}(\mathbf{r})\|^2,
\end{equation}
\begin{equation}
\mathcal{L}_{\text{TV}_{\text {coarse}}}=\frac{1}{\left|\mathbf{V}_\sigma\right|} \sum_{\mathbf{v} \in \mathbf{V}_\sigma} \sqrt{\Delta_x^2 \mathbf{v}+\Delta_y^2 \mathbf{v}+\Delta_z^2 \mathbf{v}},
\end{equation}
\begin{equation}
\mathcal{L}_\text{coarse}=\mathcal{L}_{\text {rgb}_\text {{coarse}}}+\lambda_{\text {TV}} \mathcal{L}_{\text{TV}_{\text {coarse}}},
\end{equation}
where $\lambda_{\text {TV}}=0.000016$. Here, $\mathcal{R}$ represents the set of training pixel rays, with $\mathbf{c}(\mathbf{r})$ and $\mathbf{\hat c}(\mathbf{r})$ denoting the actual and predicted colors of a ray $\mathbf{r}$, respectively. $\Delta_{x,y,z}^2\mathbf{v}$ signifies the squared difference in the voxel's density value. Notably, the total variation loss is activated only during the training iterations from 1000 to 12000.
For optimization, we utilize the Adam optimizer for training 16000 iterations with a batch size of 10192 rays. The learning rates for $\mathbf{V}_{\sigma}$, $\mathbf{V}_{c}$ and global MLP are 0.1, 0.11 and 0.002, respectively.

Once we obtain the density grid $\mathbf{V}_{\sigma}^{t}$ for each frame $t$ in the coarse training phase, we generate a per-frame occupancy grid $\mathbf{O}^{t}$ by retaining voxels with a density above the threshold $\gamma=0.003$. During our adaptive grouping stage, we set the pixel limit to $\theta=512\times 512$.

\subsection{Fine-grained Sequential Training}
After creating the mapping tables, we proceed to fine-grained sequential training within each group. 
At this stage, we also introduce a global tiny MLP $\Phi_f$ designed for efficient rendering on mobile platforms. 
This minimal MLP $\Phi_f$ consists of only one hidden layer with 16 channels, and we maintain the color feature dimension $h$ at 12. 
 Similar to the coarse stage, we progressively upscale the volume grid during training, moving from $(125\times125\times125)$ to $(150\times150\times150)$, then to $(200\times200\times200)$, and finally to $(250\times250\times250)$, corresponding to the training steps at 2000, 4000, and 6000, respectively. 
For loss calculations, we employ both the photometric MSE loss and the total variation loss on the density volume $\mathbf{V}_{\sigma}$, as well as spatial consistency loss and temporal consistency loss on the feature image $\mathbf{I}$:
\begin{equation}
\mathcal{L}_{\text {rgb}_\text {{fine}}}=\sum_{\mathbf{r} \in \mathcal{R}}\|\mathbf{c}(\mathbf{r})-\hat{\mathbf{c}}(\mathbf{r})\|^2,
\end{equation}
\begin{equation}
\mathcal{L}_{\text{TV}_{\text {fine}}}=\frac{1}{\left|\mathbf{V}_\sigma\right|} \sum_{\mathbf{v} \in \mathbf{V}_\sigma} \sqrt{\Delta_x^2 \mathbf{v}+\Delta_y^2 \mathbf{v}+\Delta_z^2 \mathbf{v}}
\end{equation}
\begin{equation}
\mathcal{L}_{\mathrm{spatial}}=\frac{1}{|\mathcal{P}|} \sum_{\substack{\mathbf{p} \in \mathcal{V}}} \left( \Delta_u(\mathbf{p})+\Delta_v(\mathbf{p}) \right),
\end{equation}
\begin{equation}
\mathcal{L}_{\mathrm{temporal}}= \|\mathbf{I}^t-\mathbf{I}^{t-1}\|_1 ,
\end{equation}
\begin{equation}
\mathcal{L}_{\text {fine}}= \mathcal{L}_{\mathrm{rgb}_\text{fine}}+\lambda_{\text {TV}} \mathcal{L}_{\text{TV}_{\text {fine}}}+\lambda_\text {s} \mathcal{L}_{\mathrm{spatial}}  +\lambda_\text {t}\mathcal{L}_{\mathrm{temporal}}, 
\end{equation}
where $\lambda_{\text {TV}}=0.000016$, $\lambda_{\text {s}}=0.0001$ and   $\lambda_{\text {t}}=0.0001$. The total variation loss is specifically activated during training iterations 1000 to 12000. We continue to use the Adam optimizer for 16000 iterations with a batch size of 10192 rays. The learning rates for $\mathbf{V}_{\sigma}$, $\mathbf{V}_{c}$, and the global MLP are set to 0.1, 0.11, and 0.002, respectively.

\begin{figure}[t]
	\begin{center}
		\includegraphics[width=0.98\linewidth]{ 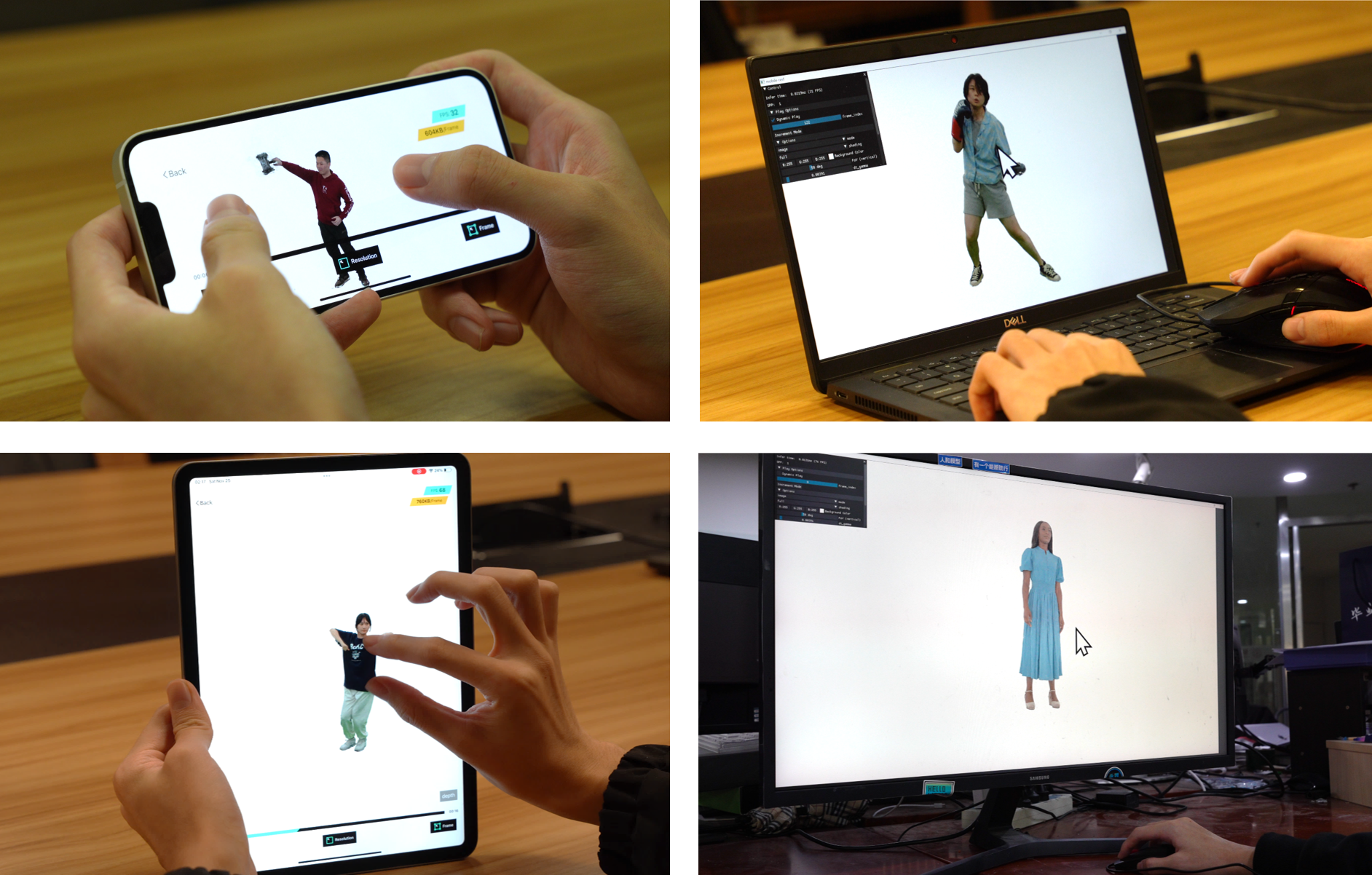}
	\end{center}
        \vspace{-5mm}
	\caption{Our VideoRF facilitates dynamic radiance field rendering on ubiquitous devices, from desktops to mobile phones.}
	\label{fig:devices}
	\vspace{-5mm}
\end{figure}
\section{Implementation Details for VideoRF Player}
\label{sec:player}

During the baking process in our training scheme, we generate a mapping table. 
For streaming, we save it in a 2D-to-3D format as an RGB image to conserve bitrate. 
Then, to enable rapid rendering, we first use this mapping table to recover a 3D volume.  
We employ a compute shader for efficient processing, segmenting the 512$\times$512 mapping table into 16$\times$16 workgroups, each handling a 32$\times$32 pixel section.  
The highly parallel architecture of compute shaders enables us to efficiently convert 2D features back into a 3D volume.
Meanwhile, we adopt a multi-resolution occupancy grid to bypass empty 3D spaces at various levels.
This approach significantly reduces unnecessary network inferences during the ray marching process. 
The largest occupancy grid is derived from max-pooling the full-resolution binary mask. 
Each subsequent grid is designed to be half the resolution of its predecessor. 
For instance, considering our full-resolution binary mask is of size 288$\times$288$\times$288, our multi-resolution occupancy grids follow suit with sizes of 144$\times$144$\times$144, 72$\times$72$\times$72, 36$\times$36$\times$36, 18$\times$18$\times$18, and 9$\times$9$\times$9. 

\section{Additional Experiments}
\begin{table}[t]
 \centering
 \resizebox{\linewidth}{!}{
\begin{tabular}{clcccccc}
\multicolumn{8}{c}{ \colorbox{best1}{best} \colorbox{best2}{second-best} } \\
\multicolumn{1}{l}{Method} & Metric & 20 & 50 & 100 & 250 & 500 & 1000 \\
\midrule
\multirow{3}{*}{HumanRF}& $\downarrow$ LPIPS & \cellcolor{best1}0.120 & \cellcolor{best1}0.138 & \cellcolor{best1}0.135 & \cellcolor{best1}0.151 & \cellcolor{best1}0.155 & \cellcolor{best1}0.160 \\
& $\uparrow$ PSNR & \cellcolor{best1}31.02 & \cellcolor{best1}30.26 & \cellcolor{best1}30.25 & \cellcolor{best1}28.98 & \cellcolor{best1}29.50 & \cellcolor{best1}29.19 \\
& $\uparrow$ SSIM & \cellcolor{best1}0.893 & \cellcolor{best1}0.888 & \cellcolor{best1}0.896 & \cellcolor{best1}0.888 & \cellcolor{best1}0.885 & \cellcolor{best1}0.881 \\
\midrule
\multirow{3}{*}{TiNeuVox}& $\downarrow$ LPIPS & 0.352 & 0.298 & 0.406 & 0.430 & 0.436 & 0.452 \\
& $\uparrow$ PSNR & 27.51 & 26.62 & 24.13 & 22.98 & 22.30 & 21.28 \\
& $\uparrow$ SSIM & 0.782 & 0.791 & 0.760 & 0.752 & 0.751 & 0.747 \\
\midrule
\multirow{3}{*}{NDVG}& $\downarrow$ LPIPS & 0.240 & 0.281 & 0.354 & 0.435 & 0.453 & 0.481 \\
& $\uparrow$ PSNR & 28.76 & 25.83 & 23.13 & 21.17 & 20.05 & 17.83 \\
& $\uparrow$ SSIM & \cellcolor{best2}0.841 & 0.812 & 0.763 & 0.731 & 0.724 & 0.692 \\
\midrule
\multirow{3}{*}{HyperNeRF}& $\downarrow$ LPIPS & \cellcolor{best2}0.233 & \cellcolor{best2}0.250 & 0.275 & 0.322 & 0.374 & 0.388 \\
& $\uparrow$ PSNR & 25.75 & 26.53 & 25.96 & 24.85 & 23.29 & 23.04 \\
& $\uparrow$ SSIM & 0.827 & 0.818 & 0.800 & 0.777 & 0.758 & 0.761 \\
\midrule
\multirow{3}{*}{NeuralBody}& $\downarrow$ LPIPS & 0.288 & 0.333 & 0.354 & 0.368 & 0.396 & 0.429 \\
& $\uparrow$ PSNR & 27.51 & 25.88 & 27.18 & 25.30 & 24.81 & 25.68 \\
& $\uparrow$ SSIM & 0.804 & 0.777 & 0.739 & 0.762 & 0.745 & 0.668 \\
\midrule
\multirow{3}{*}{TAVA}& $\downarrow$ LPIPS & 0.261 & 0.303 & 0.341 & 0.410 & 0.467 & 0.504 \\
& $\uparrow$ PSNR & 28.47 & 26.93 & 25.83 & 24.28 & 23.13 & 22.21 \\
& $\uparrow$ SSIM & 0.820 & 0.801 & 0.782 & 0.749 & 0.721 & 0.704 \\
\midrule
\multirow{3}{*}{MeRF}& $\downarrow$ LPIPS & 0.278 & 0.276 & \cellcolor{best2}0.259 & \cellcolor{best2}0.271 & \cellcolor{best2}0.272 & \cellcolor{best2}0.263 \\
& $\uparrow$ PSNR & 28.24 & 28.19 & 27.24 & 27.22 & 27.31 & 27.68 \\
& $\uparrow$ SSIM & 0.783 & 0.791 & 0.815 & 0.807 & 0.805 & 0.814 \\
\midrule
\multirow{3}{*}{ReRF}& $\downarrow$ LPIPS & 0.297 & 0.296 & 0.297 & 0.296 & 0.292 & 0.294 \\
& $\uparrow$ PSNR & 28.69 & 28.51 & 28.55 & 28.33 & 28.12 & 27.73 \\
& $\uparrow$ SSIM & 0.834 & 0.828 & 0.827 & 0.836 & 0.836 & 0.841 \\
\midrule
\multirow{3}{*}{Ours}& $\downarrow$ LPIPS & 0.276 & 0.285 & 0.283 & 0.278 & 0.274 & 0.275 \\
& $\uparrow$ PSNR & \cellcolor{best2}29.14 & \cellcolor{best2}28.79 & \cellcolor{best2}28.81 & \cellcolor{best2}28.46 & \cellcolor{best2}28.50 & \cellcolor{best2}28.32 \\
& $\uparrow$ SSIM & 0.840 & \cellcolor{best2}0.835 & \cellcolor{best2}0.830 & \cellcolor{best2}0.838 & \cellcolor{best2}0.840 & \cellcolor{best2}0.844 \\
\bottomrule
\end{tabular}
}
\caption{Quantitative comparison on long-duration sequence. We evaluate on the Actor 3, Sequence 1 of the Actors-HQ Dataset.}
\label{long}
\end{table}
\begin{figure}[t]
	\begin{center}
		\includegraphics[width=\linewidth]{ 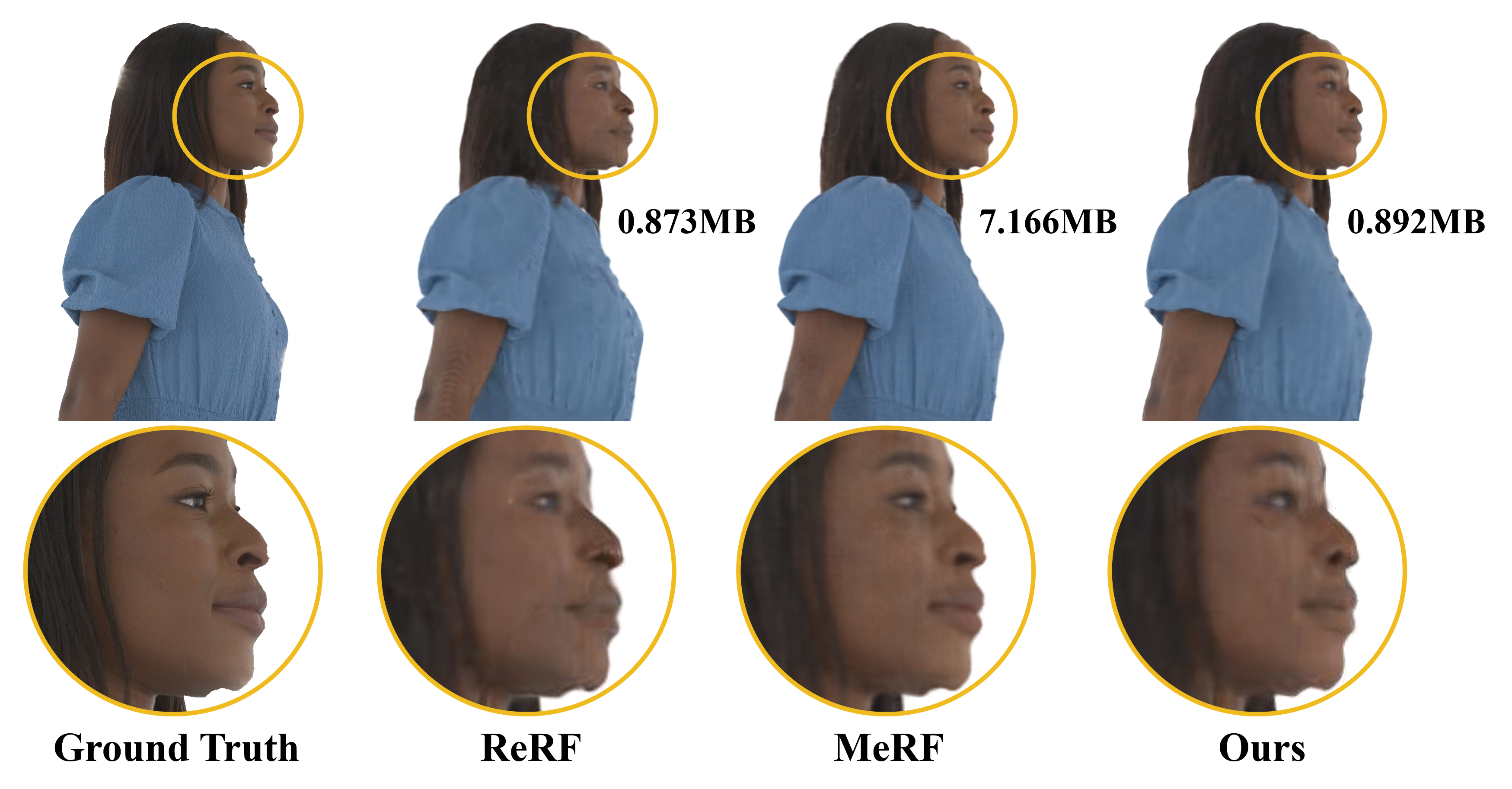}
	\end{center}
        \vspace{-5mm}
	\caption{Qualitative comparison on the long-duration sequence against recent dynamic scene reconstruction methods and per frame static reconstruction methods.}
	\label{fig:long_comp}
	\vspace{-5mm}
\end{figure}

As illustrated in Fig. \ref{fig:devices}, our method can enable dynamic radiance field rendering on a wide range of devices, including desktops (an i7-12700F
CPU and NVIDIA RTX3090 GPU), laptops (an i5-1135G7CPU and Integrated GPU), tablets (iPad Pro, an M2 chip) and mobile phones  (iPhone 14 Pro, an A16 Bionic chip).
\\
\textbf{Long-duration dynamic scenes.} Following the approach in HumanRF \cite{isik2023humanrf}, we assess performance on a long-duration sequence (Actor3, sequence1, 1000 frames) from the Actors-HQ dataset.
We compare our method with ReRF \cite{Wang_2023_rerf} and MeRF \cite{Reiser2023SIGGRAPH} through per-frame static reconstruction in Fig. \ref{fig:long_comp}.
Our method keeps a small storage to enable streaming while maintaining a high rendering quality.
We adopt the testing methods outlined in HumanRF. 
The performance metrics for HumanRF \cite{isik2023humanrf}, TiNeuVox \cite{TiNeuVox}, NDVG \cite{Guo_2022_NDVG_ACCV}, HyperNeRF \cite{park2021hypernerf}, NeuralBody \cite{peng2021neural}, TAVA \cite{li2022tava}  are directly sourced from the HumanRF publication. 
As shown in Fig. \ref{fig:long_comp} and Tab. \ref{long}, our approach demonstrates its capability to sustain high photorealism which is only second to HumanRF throughout long-duration sequences.  Note that, our VideoRF is the only method to enable rendering dynamic scenes on mobile platforms.

\end{document}